\def\PUBLICARXIV{1}
\documentclass[letterpaper]{article} 
\ifdefined\PUBLICARXIV
  \usepackage[preprint]{aaai2027-arxiv}
\else
  \usepackage[preprint]{aaai2027}
\fi
\usepackage[hyphens]{url} 
\usepackage{graphicx} 
\usepackage{natbib} 
\usepackage{caption} 
\usepackage{algorithm}
\usepackage{algorithmic}
\usepackage{booktabs}
\usepackage{array}
\usepackage{amsmath}
\usepackage{amssymb}
\usepackage{bm}
\usepackage{xcolor}
\usepackage{multirow}
\usepackage{placeins}

\definecolor{SupplementBlue}{HTML}{275D99}

\newcommand{\suppref}[2]{%
  \ifdefined\PUBLICARXIV
    \hyperref[#1]{Appendix~#2}%
  \else
    Supplementary Section~#2%
  \fi
}

\ifdefined\PUBLICARXIV
  \definecolor{PublicLinkBlue}{HTML}{275D99}
  \usepackage[
    colorlinks=true,
    linkcolor=PublicLinkBlue,
    citecolor=PublicLinkBlue,
    urlcolor=PublicLinkBlue,
    bookmarks=true,
    bookmarksopen=true,
    breaklinks=true,
    pdfstartview=FitH,
    pdftitle={When Does Muon Help Agentic Reinforcement Learning?},
    pdfauthor={Kai Ruan, Jinghao Lin, Zihe Huang, Ziqi Zhou, Qianshan Wei, Xuan Wang, and Hao Sun}
  ]{hyperref}
\fi

\title{When Does Muon Help Agentic Reinforcement Learning?}

\author{
    Kai Ruan\textsuperscript{\rm 1},
    Jinghao Lin\textsuperscript{\rm 2},
    Zihe Huang\textsuperscript{\rm 3}\\
    Ziqi Zhou\textsuperscript{\rm 4},
    Qianshan Wei\textsuperscript{\rm 5},
    Xuan Wang\textsuperscript{\rm 6},
    Hao Sun\textsuperscript{\rm 1,*}
}

\affiliations{
    \textsuperscript{\rm 1}Gaoling School of Artificial Intelligence, Renmin University of China\\
    \textsuperscript{\rm 2}Independent Researcher\\
    \textsuperscript{\rm 3}Institute of Computing Technology, Chinese Academy of Sciences\\
    \textsuperscript{\rm 4}Duke University\\
    \textsuperscript{\rm 5}Institute of Automation, Chinese Academy of Sciences\\
    \textsuperscript{\rm 6}College of Computer Science and Technology, Zhejiang University\\
    \textsuperscript{\rm *}Corresponding author
}

\begin{document}

\maketitle

\begin{abstract}
Muon is competitive with AdamW in large-scale pre-training, but its operating
regime in reinforcement-learning post-training remains unclear. We map this
regime on ALFWorld, a sparse-reward agentic benchmark, using three
group-based objectives and Qwen2.5 models from 0.5B to 3B. Under a shared KL
and clipping recipe, matched optimizer comparisons and AdamW rate controls trace the usable step-size
range. AdamW responds non-monotonically to rate, whereas fan-in Muon remains
stable at a more aggressive effective step: at $3\times10^{-5}$ it improves
late success over an AdamW $10^{-6}$ baseline after correction across
rate--metric tests. Its normalized-AUC effect is directionally positive but
less uniform; the heuristic-matched lower-rate effect is less consistent, and
tuned AdamW nearly matches high-rate Muon at 3B GraphGPO. High-rate Muon applies
$3.53\times$ AdamW's hidden-matrix update RMS; a full-budget RMS-matched control
removes the late-success gain. Together, these results identify a
recipe-level operating regime in which fan-in Muon supports a more aggressive stable effective step
under shared KL and clipping: the margin is largest when optimization headroom
remains and contracts near saturation, after AdamW tuning, or under magnitude
matching. The scale-matched control ties this spectral effect to Muon's scale
convention rather than establishing a universal optimizer ranking.
\ifdefined\PUBLICARXIV
Code is available at \href{https://github.com/x66ccff/verl-muon}{\texttt{x66ccff/verl-muon}}.
\fi
\end{abstract}

\section{Introduction}

Muon \citep{jordan2024muon} replaces the element-wise adaptive scaling of
Adam-family optimizers with an approximate spectral normalization of the
momentum matrix, computed via Newton--Schulz (NS) iterations. In
pre-training, this simple change is remarkably effective: Muon matches
AdamW's final loss with roughly $52\%$ of the training FLOPs at
billion-parameter scale \citep{liu2025muonscalable}, and has since been
used in trillion-parameter pre-training \citep{kimi2025k2}. Yet its authors
explicitly identified whether Muon transfers to \emph{post-training},
particularly reinforcement learning (RL), as an open question
\citep{jordan2024muon}.

Post-training evidence is mixed. NeMo RL reports minor gains from introducing
Muon during post-training \citep{nemorl2026muon}, whereas supervised
fine-tuning studies find that switching an Adam-pretrained model to Muon can
worsen the learning--forgetting tradeoff at poorly chosen rates
\citep{qu2026canmuon,liu2026optconsistency}. In RLVR, \citet{fan2026pion}
attribute vanilla Muon failures to spectral whitening of noise-dominated
policy-gradient directions; related engineering studies report similar
instability \citep{wei2026fieldnotes,wei2026hopper}. This evidence establishes
strong regime and learning-rate dependence, but concerns primarily
single-turn, outcome-supervised training.

We ask when Muon helps long-horizon, sparse-reward agentic RL. On ALFWorld
\citep{shridhar2020alfworld}, we train Qwen2.5-Instruct agents at 0.5B, 1.5B,
and 3B scales \citep{yang2024qwen25} with three group-based objectives: GRPO
\citep{shao2024deepseekmath}, GiGPO \citep{feng2025gigpo}, and GraphGPO
\citep{cheng2026graphgpo}. These objectives provide heterogeneous
credit-assignment conditions for testing whether an optimizer regime persists;
a targeted ablation separately tests optimizer-by-credit interaction.

\begin{figure*}[t!]
\centering
\includegraphics[width=\textwidth]{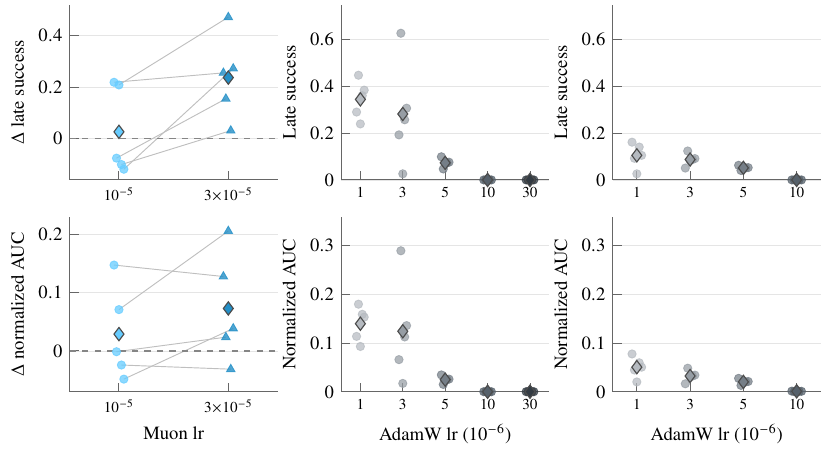}
\caption{Optimizer and rate evidence. Rows report late success and normalized
AUC. \textbf{Left:} Muon minus matched AdamW $10^{-6}$ in GiGPO
comparisons; gray segments connect Muon rates within a comparison. \textbf{Middle
and right:} GiGPO and GRPO AdamW outcomes by learning rate. Small markers show
individual outcomes and diamonds show means;
\suppref{supp:matched-robustness}{C.2} reports the individual comparisons.}
\label{fig:gigpo-lr-sweep}
\end{figure*}

Figure~\ref{fig:gigpo-lr-sweep} frames the central comparison as an operating-
regime question. The matched comparisons measure Muon against an AdamW baseline,
while separate rate controls show when that baseline is
competitive. AdamW at $3\times10^{-6}$ is competitive but variable under
GiGPO; under GRPO, $10^{-6}$ retains the strongest tested mean. Across the
matched comparisons, the pooled paired evidence supports a high-rate
Muon advantage in late success; the normalized-AUC effect is also directionally
positive after correction, but its estimator-specific magnitude is less uniform.
A 3B GraphGPO
comparison provides a boundary case: high-rate Muon learns earlier, while
tuned AdamW and high-rate Muon have nearly identical AUC.
The lower Muon rate, which follows the practical $10\times$ conversion
heuristic, is not consistently beneficial. The strongest evidence therefore
concerns the stable operating range of the aggressive fan-in recipe.

Scale changes the visible form of improvement. At 1.5B, both Muon rates attain
higher AUC than AdamW across the evaluated estimators, while late success
depends on remaining headroom. At 3B GraphGPO, tuning AdamW nearly closes the
AUC gap; in saturated GiGPO settings, high-rate Muon reaches the strong-policy
regime earlier. Applied-update measurements show that the optimizer choice
changes both step magnitude and spectral allocation. A matched probe measures
the scale difference at
$3.53\times$ AdamW's hidden-matrix RMS, and a full-budget RMS-matched control
loses the standard high-rate recipe's advantage.
A targeted credit-control ablation finds strong main effects and limited
optimizer-by-credit interaction. We therefore use the spectral measurements
as diagnostics and summarize a credit-conditioned interpretation as a compact
hypothesis in \suppref{supp:credit-hypothesis}{D}.

Our contributions are:
\begin{itemize}
    \item Controlled optimizer comparisons with explicit AdamW rate sweeps identify
    the operating regimes of fixed fan-in Muon and AdamW recipes under a shared
    KL and clipping configuration.
    \item Scale and transfer extensions separate early progress from final quality:
    aggressive Muon is most useful when optimization headroom remains, while
    AdamW tuning and task saturation narrow the difference.
    \item Applied-update measurements and a full-budget RMS-matched control
    identify the larger applied step as a material component of the observed
    gain and directly characterize Muon's spectral redistribution.
\end{itemize}

\section{Related Work}

\paragraph{Muon and matrix-aware optimizers.}
Muon \citep{jordan2024muon} orthogonalizes the momentum matrix of each 2D
hidden-layer parameter via Newton--Schulz iterations, and is typically
deployed with an auxiliary Adam-family optimizer for embeddings, norms,
and output heads. \citet{liu2025muonscalable} demonstrated scalability to
LLM pre-training with roughly half of AdamW's FLOP budget. MuonClip
\citep{kimi2025k2} stabilized trillion-parameter pre-training by controlling
attention logits, a different failure mode from the RL behavior studied here.
With decoupled weight decay, Muon is a nuclear-norm Lion-$\mathcal{K}$ method
that implicitly
enforces a spectral-norm constraint, although this characterization does not
predict RL behavior \citep{chen2025muonconstraints}. HTMuon relaxes uniform
orthogonalization with a heavy-tailed singular-value transform, motivated by
the risk of overweighting noise-dominated directions; its evidence comes from
pre-training and vision rather than RL \citep{pang2026htmuon}. Pion makes a
related diagnosis in RLVR and replaces uniform spectral whitening with a
high-pass NS transformation that suppresses tail components
\citep{fan2026pion}. Our work instead keeps the Keller--Jordan Muon
transformation fixed while varying the advantage estimator.

Concurrent Muon variants modify complementary parts of this transformation.
Muon$^2$ preconditions the momentum with an adaptive second moment before NS
\citep{liu2026muon2}; AMO allocates the NS budget by operator geometry
\citep{zhuang2026amo}; Muown controls row magnitude to reduce spectral-norm
drift and rate sensitivity \citep{lion2026muown}; and MiMuon mixes
orthogonalized and momentum updates \citep{huang2026mimuon}. Their evidence
concerns pre-training or supervised generalization rather than long-horizon RL,
so we treat them as complementary optimizer designs rather than evaluated
baselines.

Recent work sharpens two aspects that are central to our design. Update-RMS
matching and layer-wise distributed implementations extend SOAP and Muon to
multi-billion-parameter pre-training \citep{khona2026soap}; NAMO and NorMuon
combine orthogonalization with adaptive global or neuron-wise scaling
\citep{zhang2026namo,li2025normuon}. A complementary analysis proves
condition-number-independent convergence for simplified Muon in matrix
factorization and linear-transformer in-context learning
\citep{ma2026preconditioning}. These results motivate our update-scale controls
and spectral diagnostics, but do not determine the stochastic policy-optimization
regime studied here.

\paragraph{Muon in post-training and optimizer mismatch.}
\citet{qu2026canmuon} and \citet{liu2026optconsistency} study optimizer
mismatch in supervised fine-tuning. Both find that switching an Adam-pretrained
model to Muon is sensitive to learning rate and can worsen the
learning--forgetting tradeoff, whereas optimizer continuity is more favorable.
Muon-pretrained models have also been successfully instruction-tuned with Muon
\citep{gupta2025muonquantization}, and large Muon-pretrained models have
subsequently undergone RL stages \citep{kimi2025k2,primeintellect2025intellect3}.
For RLVR itself, \citet{fan2026pion} attribute vanilla Muon failures to spectral
whitening under low gradient SNR and propose a high-pass remedy; Hopper instead
combines variance normalization with a single NS step
\citep{wei2026fieldnotes,wei2026hopper}. Together, these studies establish
strong optimizer- and rate-dependence but leave long-horizon agentic RL and its
behavior across group-based training objectives largely unexplored.

\paragraph{Group-based RL for LLM agents.}
Group-based methods, including RLOO \citep{kool2019buy,
ahmadian2024back}, GRPO \citep{shao2024deepseekmath}, and DAPO
\citep{yu2025dapo}, estimate advantages from within-group statistics of
rollouts. For multi-turn agents,
credit assignment across long horizons is the central difficulty. GiGPO
\citep{feng2025gigpo} adds a step-level relative advantage by retroactively
grouping actions taken from repeated anchor states across trajectories,
at negligible cost and without extra rollouts. GraphGPO
\citep{cheng2026graphgpo} aggregates rollouts into a unified state-transition
graph, estimates each state's distance to the task goal, and assigns
edge-level advantages from successor-state distance; under deterministic-
environment and fixed-policy assumptions, it also gives a conditional
variance result for scalar feedback. The three methods therefore provide
distinct episode-, step-, and transition-level credit constructions with which
to test the breadth of an optimizer regime; the GiGPO $\omega$ control isolates
one credit interaction directly.

HGPO instead addresses context inconsistency within step-level groups by
constructing a hierarchy of history-conditioned groups and adaptively combining
their advantages \citep{he2026hgpo}. It is evaluated on ALFWorld and WebShop,
making it closely related on the credit-assignment axis; the present fixed
estimator sweep does not evaluate it.

Recent agent-training methods improve the signal presented to the optimizer
along complementary axes. SDAR and CRAFT introduce token-level self-distillation
and counterfactual sibling credit \citep{lu2026sdar,meng2026craft}; SCPO
recovers locally useful progress from failed trajectories
\citep{xu2026scpo}. StraTA and HiPER impose trajectory- or subgoal-level
hierarchies \citep{xue2026strata,peng2026hiper}, while EnvRL adds state and
inverse-dynamics objectives \citep{wang2026envrl}. These methods redesign the
training signal; our study instead holds the objective fixed within each
comparison and maps the optimizer recipe applied to that signal. Their
interaction with matrix-aware optimization is an open empirical direction.

\section{Background}

\paragraph{Credit structure.}
An agent emits textual actions over a horizon $T$ and receives a sparse
terminal reward. The evaluated objectives share a group-relative policy loss
but distribute this reward differently. GRPO normalizes trajectory returns,
GiGPO adds a contrast among actions sampled from repeated anchor states, and
GraphGPO derives transition credit from distance-to-goal structure in a
rollout graph \citep{shao2024deepseekmath,feng2025gigpo,cheng2026graphgpo}.
GiGPO exposes the step contribution explicitly:
\begin{equation}
A(\bm{a}_t^{(i)})=A^{E}(\bm{\tau}_i)+\omega\, A^{S}(\bm{a}_t^{(i)}),
\label{eq:gigpo-adv}
\end{equation}
where $A^E$ and $A^S$ are the episode- and anchor-state advantages. Setting
$\omega=0$ removes the step term while preserving the GiGPO code path. This
provides a targeted credit control; the primary comparisons otherwise hold
the estimator fixed and change only the policy optimizer recipe.

\paragraph{Muon.}
For each hidden weight matrix
$W\in\mathbb{R}^{d_{\mathrm{out}}\times d_{\mathrm{in}}}$ with
momentum-accumulated gradient $M=U\Sigma V^{\top}$, Muon
\citep{jordan2024muon} replaces the singular-value magnitudes of $M$ by an
approximate polar factor $Q=\mathrm{NS}_5(M)\approx UV^{\top}$. The
Keller--Jordan implementation used in our experiments applies the
fan-in-scaled update
\begin{equation}
\Delta W=-\eta_{\mathrm{KJ}}
\sqrt{\max\!\left(1,\frac{d_{\mathrm{out}}}{d_{\mathrm{in}}}\right)}\,Q.
\label{eq:muon-update-scaling}
\end{equation}
Non-matrix parameters are updated with AdamW; implementation and partitioning
details appear in \suppref{supp:update-scale}{B}. For an ideal polar
factor, this gives
$\operatorname{RMS}(\Delta W)\approx
\eta_{\mathrm{KJ}}/\sqrt{d_{\mathrm{in}}}$ before weight decay.

\paragraph{Learning-rate comparability.}
Muon learning rates depend on the matrix-shape convention. Relating the
fan-in-scaled update to the RMS-matched convention of
\citet{liu2025muonscalable}, with $c\approx0.2$, gives
\begin{equation}
\eta_{\mathrm{RMS}}
\approx
\frac{\eta_{\mathrm{KJ}}}{c\sqrt{d_{\mathrm{in}}}},
\label{eq:muon-lr-conversion}
\end{equation}
Across our matrix shapes, this conversion agrees with the practical
$10\times$ heuristic of \citet{su2025muonguide}. We therefore pair AdamW at
$10^{-6}$ with Muon at $10^{-5}$ and $3\times10^{-5}$: the lower Muon rate
follows that heuristic, while the higher rate probes a more aggressive fan-in
recipe and is not presented as rate-equivalent to AdamW $10^{-6}$. Fixing both
rates across estimators and objectives separates the paired
comparison from post-hoc rate selection; a separate AdamW sweep measures rate
sensitivity. The main study therefore compares fixed optimizer
recipes over the tested range. Matched diagnostics and a full-budget control in
\suppref{supp:update-scale}{B} measure applied-update scale and evaluate an
RMS-matched Muon convention.

\section{Experiments}

\subsection{Experimental Setup}

\paragraph{Environment and scale.}
ALFWorld \citep{shridhar2020alfworld} comprises long-horizon embodied household
tasks with sparse terminal success and an invalid-action penalty.
We additionally retain the 0.5B GiGPO recipe for a transfer case on
WebShop \citep{yao2022webshop}, a simulated e-commerce environment with
continuous partial task scores and exact task-completion outcomes.
The estimator comparison uses Qwen2.5-0.5B-Instruct. Scale extensions use
Qwen2.5-1.5B-Instruct under all three estimators and Qwen2.5-3B-Instruct under
GiGPO and GraphGPO \citep{yang2024qwen25}.
Training uses verl-agent \citep{feng2025gigpo}, an agentic extension of
veRL/HybridFlow \citep{sheng2024hybridflow}, with vLLM rollouts
\citep{kwon2023vllm}. Training and evaluation schedules are held fixed across
the matched comparisons. The complete training configuration appears in
\suppref{supp:hparams}{A}, with optimizer-partition and numerical details in
\suppref{supp:update-scale}{B}.

\paragraph{Comparison design.}
At 0.5B, we cross the three estimators with an AdamW policy baseline at
$10^{-6}$ and hidden-matrix Muon at $10^{-5}$ or $3\times10^{-5}$; fallback
parameters retain AdamW at $10^{-6}$. Within each matched comparison, we hold
the model, data, rollout protocol, estimator, training schedule, environment
initialization, task split, evaluation set, and all non-optimizer settings
fixed. Unless otherwise stated, matched cells use five distinct random seeds.
Separate AdamW sweeps measure rate sensitivity.
Targeted controls set $\omega=0$ under GiGPO. GRPO rate
controls and default- and tuned-rate 3B GraphGPO comparisons appear in
\suppref{supp:rate-controls}{E.1}. Scale tests cover all three objectives at
1.5B and GiGPO and GraphGPO at 3B. The WebShop test retains the 0.5B GiGPO
comparison and training schedule while changing the environment.

\paragraph{Metrics.}
We summarize each training trajectory by late-window validation success and
normalized validation AUC over the full checkpoint curve. The former measures
late-training quality; the latter captures progress over the fixed update
budget when final success saturates. AUC measures progress per scheduled policy
update; wall-clock measurements are reported separately. Rollout-policy entropy
is reported as a training diagnostic.
We report variation across matched comparisons. The pooled sign test
provides an omnibus direction check across heterogeneous estimator strata;
estimator-specific means and dispersions provide the effect-size summaries.
We Holm-correct the rate--metric family.
For WebShop, the same late-window and trapezoidal-AUC summaries are applied to
its continuous task score; exact success is reported separately because task
completion is rare in this setting.

\subsection{Learning-Rate Controls}

Figure~\ref{fig:gigpo-lr-sweep} is organized by metric in rows and comparison
in columns. The left column reports Muon minus matched AdamW $10^{-6}$ for
GiGPO. The middle and right columns map AdamW's response over four
learning rates under GiGPO and GRPO, without pooling unmatched configurations
into a single optimizer ranking.

Across the matched GiGPO comparisons, high-rate Muon improves late success,
while its AUC effect varies across comparisons. The AdamW controls reveal a
non-monotonic response: the intermediate rate is competitive but variable and
does not consistently improve on the $10^{-6}$ reference. Rates of
$5\times10^{-6}$ and above lose nearly all
post-update success. Under GRPO, $10^{-6}$ retains the strongest tested mean,
whereas $10^{-5}$ loses post-update success.
The paired comparison preserves configuration matching; the sweeps expose how the
AdamW baseline depends on optimizer rate. All other AdamW settings, including
momentum, warmup, and gradient clipping, remain fixed. Exact rate summaries,
cross-estimator policy movement, and high-rate stress trajectories appear in
\suppref{supp:rate-controls}{E.1}, \suppref{supp:optimizer-dynamics}{C.4}, and
\suppref{supp:failure-diagnostics}{E.2}.
These controls map the stable rate range under the shared KL and clipping
recipe; joint rate--regularization tuning may move that boundary.

\subsection{Paired Effects Across Objectives}

\begin{figure*}[t!]
\centering
\includegraphics[width=\textwidth]{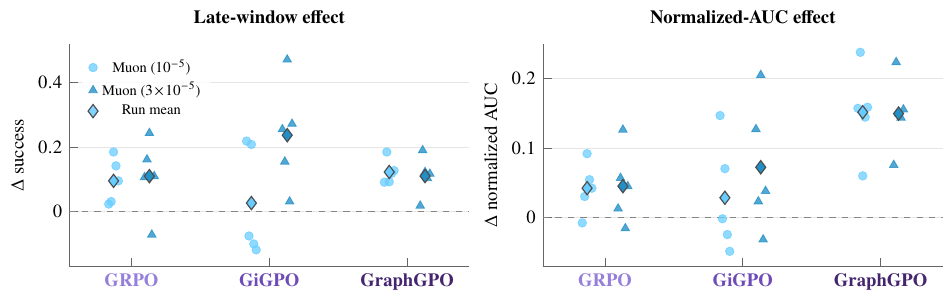}
\caption{Paired optimizer effects at 0.5B. Each point is Muon minus its
matched AdamW $10^{-6}$ baseline; diamonds show
the mean within each estimator. This figure estimates the effect relative to
the AdamW baseline, while Figure~\ref{fig:gigpo-lr-sweep} reports
AdamW rate sensitivity.}
\label{fig:algofamily}
\end{figure*}

Figure~\ref{fig:algofamily} and Table~\ref{tab:main} summarize the matched 0.5B
optimizer comparisons. The high-rate Muon recipe produces a broadly positive
late-success effect and a less uniform normalized-AUC effect. Exact two-sided
sign tests across the matched comparisons
give raw $p=.001$ and $p=.007$, respectively. Holm adjustment across
the four rate--metric tests gives $p_{\mathrm{adj}}=.004$ for late success and
$p_{\mathrm{adj}}=.022$ for normalized AUC. The lower-rate recipe is not
consistently signed. These pooled tests summarize direction across
heterogeneous strata. We treat late success as the primary robust endpoint;
the AUC result is a secondary progress measure whose effect sizes vary more
across estimators. Table~\ref{tab:main} reports estimator-specific
effect sizes, with the largest variation under GiGPO. Estimator-specific
intervals and sensitivity analyses appear in \suppref{supp:matched-robustness}{C.2}.

\begin{table*}[t]
\centering
\small
\setlength{\tabcolsep}{5.5pt}
\begin{tabular}{@{}lcccc@{}}
\toprule
& \multicolumn{2}{c}{Late-success difference} &
\multicolumn{2}{c}{Normalized-AUC difference} \\
\cmidrule(lr){2-3}\cmidrule(l){4-5}
Estimator & Muon $10^{-5}$ & Muon $3{\times}10^{-5}$ &
Muon $10^{-5}$ & Muon $3{\times}10^{-5}$ \\
\midrule
GRPO     & $+0.095\pm0.070$ & $+0.110\pm0.116$ & $+0.043\pm0.036$ & $+0.045\pm0.053$ \\
GiGPO    & $+0.027\pm0.171$ & $+0.237\pm0.162$ & $+0.029\pm0.080$ & $+0.073\pm0.093$ \\
GraphGPO & $+0.123\pm0.038$ & $+0.110\pm0.061$ & $+0.152\pm0.063$ & $+0.150\pm0.052$ \\
\bottomrule
\end{tabular}
\caption{Paired Muon-minus-AdamW effects at 0.5B (mean $\pm$ sample standard
deviation across matched optimizer comparisons). AdamW uses the $10^{-6}$
baseline. Figure~\ref{fig:gigpo-lr-sweep}
provides the corresponding learning-rate controls.}
\label{tab:main}
\end{table*}

The paired analysis estimates the high-rate late-success effect relative to
the AdamW baseline. GiGPO also contains a competitive $3\times10^{-6}$ AdamW
outcome, while the GRPO sweep leaves the reference-rate mean strongest. Training
trajectories and task-category breakdowns appear in
\suppref{supp:matched-robustness}{C.2} and
\suppref{supp:task-results}{C.3}.
The same supplementary section reports 95\% $t$ intervals for each
estimator--rate effect and the estimator-stratified permutation sensitivity.

\begin{figure}[H]
\centering
\includegraphics[width=\columnwidth]{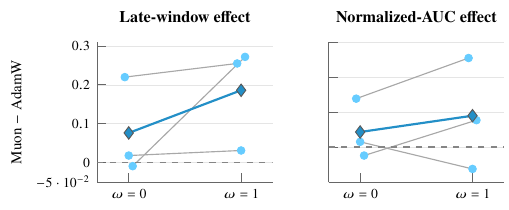}
\caption{Matched GiGPO optimizer-by-credit ablation across factorial
comparisons. Each thin line connects paired outcomes across credit
settings; diamonds and blue lines show the mean across comparisons.}
\label{fig:omega-factorial}
\end{figure}

\paragraph{Credit-control ablation.}
Matched $2\times2$ comparisons separate the optimizer effect from the
GiGPO step term. Figure~\ref{fig:omega-factorial} reports the paired effects.
Enabling the step term raises late success for both optimizers across the
comparisons. Muon improves late success under $\omega=1$, whereas the
$\omega=0$ effect is smaller and can reverse slightly. The resulting
optimizer-by-credit interaction is limited over the evaluated range.
\suppref{supp:credit-control}{C.1} provides the individual trajectories and
complete factorial summary.

\subsection{Scale Separates Early Progress from Final Quality}

\begin{figure*}[t!]
\centering
\includegraphics[width=\textwidth]{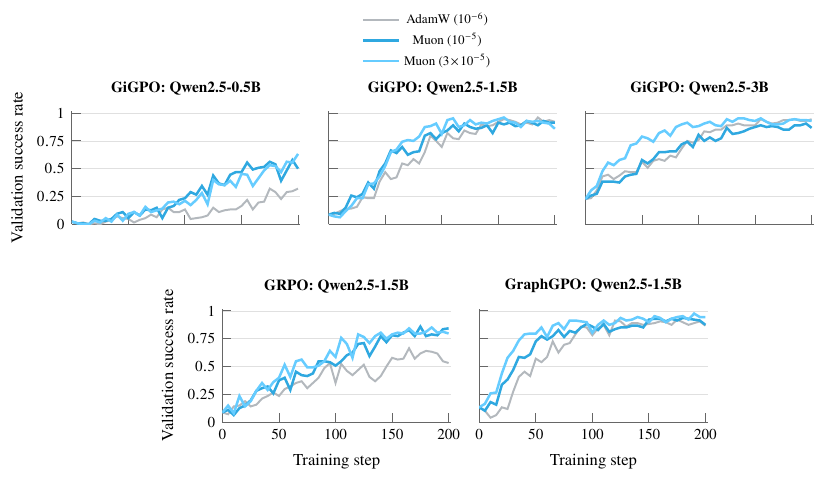}
\caption{Scale extensions under configuration-matched optimizer comparisons.
\textbf{Top:} GiGPO across 0.5B, 1.5B, and 3B. \textbf{Bottom:} GRPO and
GraphGPO at 1.5B; the top-center GiGPO panel completes the estimator
comparison. Saturated GiGPO comparisons emphasize earlier progress; GRPO and
GraphGPO retain late-success headroom at 1.5B.}
\label{fig:scale}
\end{figure*}

Figure~\ref{fig:scale} separates estimator and scale effects. In the GiGPO
scale sweep, high-rate Muon raises 0.5B late success from $0.290$ to $0.546$.
At 1.5B and 3B, GiGPO late success approaches saturation, so the separation
moves to learning progress: high-rate Muon raises normalized AUC from $0.635$
to $0.708$ at 1.5B and from $0.707$ to $0.805$ at 3B. The 1.5B estimator sweep
shows the complementary case: GRPO retains substantial late-success headroom,
GiGPO is saturated, and GraphGPO improves in both metrics.

Figure~\ref{fig:scale} makes the headroom distinction explicit. At 3B
GraphGPO (Table~\ref{tab:graphgpo-q3-tuned}), high-rate Muon reaches $0.75$
success first, while tuned AdamW $3\times10^{-6}$ attains nearly the same AUC.
\suppref{supp:scale-extensions}{C.5} tabulates the 1.5B outcomes and gives
per-task trajectories and probes.

\begin{table}[t]
\centering
\small
\setlength{\tabcolsep}{5pt}
\begin{tabular}{@{}lccc@{}}
\toprule
Configuration & Late & Norm. AUC & First eval $\geq0.75$ \\
\midrule
AdamW ($10^{-6}$) & \textbf{0.945} & 0.792 & 65 \\
AdamW ($3{\times}10^{-6}$) & 0.940 & 0.853 & 40 \\
Muon ($10^{-5}$) & 0.930 & 0.804 & 60 \\
Muon ($3{\times}10^{-5}$) & 0.936 & \textbf{0.856} & 30 \\
\bottomrule
\end{tabular}
\caption{3B GraphGPO boundary case. High-rate Muon reaches strong success
earlier than default AdamW; tuning AdamW largely closes the AUC gap. The last
column gives the first scheduled validation at or above $0.75$.}
\label{tab:graphgpo-q3-tuned}
\end{table}

\subsection{Cross-Environment Transfer Case}

We retain the 0.5B GiGPO recipe and replace ALFWorld with WebShop as a transfer
case for the same recipe-level pattern in a distinct long-horizon environment.
High-rate Muon raises partial-task-score AUC from $0.078$ to
$0.548$ and late score from $0.197$ to $0.854$; lower-rate Muon reaches
$0.256$ AUC and $0.554$ late score. Exact completion follows the same pattern:
high-rate Muon raises late success from $0.008$ to $0.689$ and success AUC
from $0.005$ to $0.305$. \suppref{supp:webshop}{C.6} gives the trajectories
and complete summary.

\subsection{Update-Spectrum Diagnostics}

\begin{figure}[H]
\centering
\includegraphics[width=0.94\columnwidth]{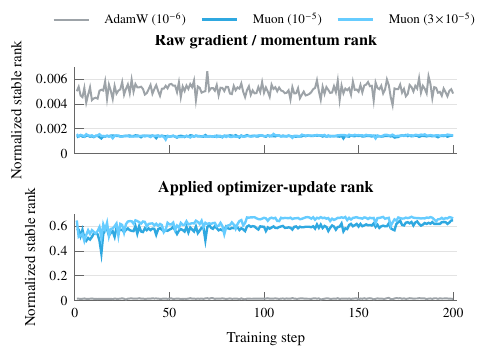}
\caption{Spectral diagnostic for a matched 1.5B GRPO comparison. The
pre-transformation first-moment direction remains low-rank for both optimizers, whereas Muon
produces a substantially flatter applied-update spectrum. This verifies the
spectral transformation independently of the task-level outcomes.}
\label{fig:main-spectral}
\end{figure}

Figure~\ref{fig:main-spectral} compares the pre-transformation first moment
with the update actually applied to a fixed set of hidden matrices. For a
probed matrix $X\in\mathbb R^{m\times n}$, normalized stable rank
$\bar r_{\mathrm{st}}(X)=\lVert X\rVert_F^2/
(\lVert X\rVert_2^2\min\{m,n\})$ increases as the spectrum flattens. A value
near zero means that a few singular directions dominate the update; a value
near one means that energy is spread nearly uniformly across available
directions. In the
matched 1.5B GRPO comparison, this statistic is $0.015$ for AdamW's applied
update and $0.585$--$0.635$ for Muon, while the raw statistic remains low for
both. Statistics are averaged equally across matrices over training.
A matched update-scale probe finds that standard high-rate Muon applies $3.53\times$ the
hidden-matrix update RMS of AdamW; an RMS-matched convention reduces this ratio
to $0.80$. At the full training budget, the RMS-matched control reaches late success
$0.255$ and normalized AUC $0.096$, below AdamW ($0.384$, $0.153$) and
standard high-rate Muon ($0.656$, $0.192$) in the matched GiGPO control.
The gain is therefore not preserved under the RMS-matched convention. Because
this convention also changes layerwise scale allocation, the experiment
identifies the scaling rule as part of the effective recipe without isolating
spectral shape.
\suppref{supp:update-scale}{B} gives scale and timing controls;
\suppref{supp:applied-spectrum}{C.7} and
\suppref{supp:scale-extensions}{C.5} give complete spectral probes.

Together, the measurements distinguish update scale from spectral shape at
the optimizer level. They establish the transformation and show that the
RMS-matched convention does not preserve the task-level gain. Because this
convention changes both global magnitude and layerwise scale allocation, the
control identifies the scaling rule as a component of the effective recipe
without isolating spectral shape. A simplified diagnostic hypothesis motivated
by these measurements appears in \suppref{supp:credit-hypothesis}{D}.

\FloatBarrier

\section{Discussion}

The combined evidence supports a concrete operating rule: fan-in Muon is most
useful when optimization headroom remains and its aggressive update is stable
under the shared KL and clipping configuration.
At 0.5B this regime improves late success; near saturation, the difference
appears as earlier progress. The lower, approximately rate-equivalent Muon
setting is inconsistent, and tuned AdamW closes the AUC gap at 3B GraphGPO.
The matched WebShop comparison provides a second-environment transfer case. In the
targeted $\omega$ ablation, the credit term improves both optimizers and its
interaction with optimizer choice is limited. The operating regime therefore
persists across all three evaluated estimators rather than depending on one
credit construction. Its rate dependence is consistent with failures reported
in single-turn RLVR
\citep{fan2026pion,wei2026fieldnotes}.

The AdamW sweep clarifies the comparison. At $3\times10^{-6}$, AdamW is
competitive but variable; higher rates fail under the shared recipe. The
paired comparison measures replacement of a fixed AdamW recipe by a fixed Muon
recipe, while the broader sweep maps AdamW's own rate sensitivity. Under the
shared KL and clipping settings, Muon sustains an effective step that AdamW
rate increases do not in the tested sweeps; joint rate--regularization tuning
may shift this boundary.

\paragraph{Practical implication.}
The results favor recipe-level rate screening over a universal conversion rule.
Under the evaluated regularization, $10^{-5}$ is the conservative Muon
candidate and $3\times10^{-5}$ is the high-headroom candidate. The high-rate
recipe should be selected only when early validation improves without
instability in response clipping, entropy, or valid-action diagnostics. Near
saturation, AdamW tuning can recover most of the difference, whereas a stable
high-rate Muon trajectory can accelerate learning under a limited update
budget.

The matched 0.5B ALFWorld comparisons provide the primary statistical evidence;
the scale and WebShop experiments are extension cases that probe whether the
observed regime persists across model size and environment. Because Muon is
applied to hidden matrices, its parameter coverage increases with model scale.
The scale results therefore characterize the full optimizer recipe rather than
model size in isolation.

The optimizer effect combines a larger stable update scale with a different
spectral allocation. The full-budget RMS-matched control falls below standard
Muon and AdamW, showing that the scaling convention is part of the effective recipe.
At this operating point, spectral flattening at an AdamW-like global magnitude
does not preserve the gain; the accompanying change in layerwise allocation
precludes isolating the spectral effect. The same measurements motivate the
diagnostic credit-reliability hypothesis in \suppref{supp:credit-hypothesis}{D}.
AUC measures progress per policy
update; matched timing shows negligible end-to-end overhead. Our conclusions
concern the shared regularization and fixed training budget; broader
checkpoint families and longer-horizon training are outside this evaluation.

\section{Conclusion}

Fan-in Muon supports a larger stable effective step in the evaluated
agent-training recipes. Under the shared regularization settings, its aggressive
setting remains stable
where larger AdamW rates are unreliable and improves paired late success; the
lower setting is inconsistent, and AdamW tuning narrows the difference.
The RMS-matched convention removes the gain, while spectral
flattening remains directly observed. The practical object is therefore the
complete optimizer recipe: scale convention, learning rate, and regularization.

\ifdefined\PUBLICARXIV
\paragraph{Acknowledgments.} We thank
\href{https://chen-yu-zheng.github.io/}{Chenyu Zheng} for helpful discussions.
\fi

\begingroup
\ifdefined\PUBLICARXIV
\small
\setlength{\bibsep}{0.5pt plus 0.2ex}
\fi
\bibliography{references}
\endgroup

\clearpage
\onecolumn
\appendix
\raggedbottom

\section{A. Experimental Hyperparameters}
\label{supp:hparams}

Table~\ref{tab:hparams} summarizes the shared setup and the
estimator-specific settings used in the reported comparisons.

\begin{table}[h]
\centering
\begingroup
\footnotesize
\setlength{\tabcolsep}{4pt}
\renewcommand{\arraystretch}{1.08}
\begin{tabular}{@{}>{\raggedright\arraybackslash}p{0.39\textwidth}%
                    >{\raggedright\arraybackslash}p{0.47\textwidth}@{}}
\toprule
Parameter & Value \\
\midrule
\multicolumn{2}{@{}l}{\textit{Shared setup}} \\
Models & Qwen2.5-Instruct (0.5B, 1.5B, 3B) \\
Environments & ALFWorld; WebShop \\
Training updates / evaluation interval & 200 / 5 \\
Max steps (ALFWorld / WebShop) & 50 / 15 \\
History length & 2 \\
Group / train / validation batch sizes & 8 / 16 / 128 \\
Prompt / response lengths & 2048 / 512 \\
PPO mini-batch / epochs / dynamic batching & 128 / 1 / yes \\
Rollout / reference micro-batch per GPU & 32 / 32 \\
Rollout / validation temperatures & 1.0 / 0.4 (sampling) \\
Hardware & $8\times$ NVIDIA H20 \\
\midrule
\multicolumn{2}{@{}l}{\textit{Optimizer and loss}} \\
Optimizer assignment & Keller--Jordan Muon for hidden 2D matrices; AdamW otherwise \\
AdamW lr & $1\times10^{-6}$ \\
Muon momentum / Nesterov / NS steps & 0.95 / yes / 5 \\
Muon NS dtype / scaling & bfloat16 / fan-in \\
Weight decay & 0.01 \\
KL loss coefficient / type / KL in reward & 0.01 / low-var KL / no \\
Invalid-action penalty (ALFWorld / WebShop) & 0.01 / 0.1 \\
\midrule
\multicolumn{2}{@{}l}{\textit{Estimator-specific settings}} \\
GRPO & Muon lr $1\times10^{-5},\ 3\times10^{-5}$ \\
GiGPO & mean-std norm; step weight $\omega=1$ (main), $0$ (ablation); Muon lr $1\times10^{-5},\ 3\times10^{-5}$ \\
GraphGPO & $\rho=0.10$; mean-std norm; step / episode weights $1/1$; Muon lr $1\times10^{-5},\ 3\times10^{-5}$ \\
GraphGPO distance normalization / similarity & disabled / disabled \\
\bottomrule
\end{tabular}
\endgroup
\caption{Core training configuration. Implementation details are reported
separately in Appendix~B.}
\label{tab:hparams}
\end{table}

\ifdefined\SUPPLEMENTARYDOC
\clearpage
\fi
\section{B. Implementation and Update-Scale Details}
\label{supp:update-scale}

\paragraph{Parameter groups.}
The optimizer uses the Keller--Jordan Muon implementation rather than an
RMS-matched or distributed variant. Hidden attention and MLP matrices are assigned to
Muon; embeddings, normalization parameters, the tied embedding/LM-head weight,
and other non-matrix parameters use AdamW. Under this rule, hidden matrices
account for approximately $72\%$, $85\%$, and $90\%$ of optimized parameters
at 0.5B, 1.5B, and 3B, respectively. Muon and fallback AdamW use separate
learning rates, while both receive the configured decoupled weight decay of
$0.01$. Newton--Schulz acts on complete 2D matrices, so the policy is
left unsharded during the optimizer step. This preserves the intended matrix
grouping and fan-in scaling; a distributed implementation would require an
equivalent full-matrix transformation.

\paragraph{Numerical kernel.}
The momentum buffer uses coefficient $0.95$ with Nesterov momentum.
Newton--Schulz orthogonalization is evaluated in bfloat16 for five iterations,
after which the update is multiplied
by $\sqrt{\max(1,d_{\mathrm{out}}/d_{\mathrm{in}})}$. This is the
Keller--Jordan fan-in convention; it
is not numerically equivalent, at the same scalar learning rate, to the
constant-update-RMS convention used by scalable Muon variants
\citep{liu2025muonscalable}.

Table~\ref{tab:muon-effective-rms} gives the resulting idealized update scale
for the two matrix shapes in the 0.5B model.

\begin{table}[h]
\centering
\small
\setlength{\tabcolsep}{6pt}
\begin{tabular}{@{}lccc@{}}
\toprule
Matrix group & $d_{\mathrm{in}}$ &
$\operatorname{RMS}(\Delta W)$ at $10^{-5}$ &
$\operatorname{RMS}(\Delta W)$ at $3{\times}10^{-5}$ \\
\midrule
Attention; MLP gate/up & 896  & $3.34{\times}10^{-7}$ & $1.00{\times}10^{-6}$ \\
MLP down projection    & 4864 & $1.43{\times}10^{-7}$ & $4.30{\times}10^{-7}$ \\
\bottomrule
\end{tabular}
\caption{Idealized per-coordinate Muon update RMS
$\eta_{\mathrm{KJ}}/\sqrt{d_{\mathrm{in}}}$, before weight decay. These values
follow analytically from the matrix shapes and scaling convention.}
\label{tab:muon-effective-rms}
\end{table}

\paragraph{Matched applied-update probe.}
A five-update GiGPO diagnostic isolates update scale with identical data,
rollout, model, initialization, and hardware settings. Alongside AdamW and the standard
fan-in Muon recipe, we evaluate the RMS-matched convention of
\citet{liu2025muonscalable}, which multiplies the base rate by
$0.2\sqrt{\max(d_{\mathrm{out}},d_{\mathrm{in}})}$ for each hidden matrix.

\begin{table}[h]
\centering
\small
\setlength{\tabcolsep}{5pt}
\begin{tabular}{@{}lrrrr@{}}
\toprule
Recipe & Hidden RMS & Total RMS & Opt. ms & Train s \\
\midrule
AdamW ($10^{-6}$) & $2.26{\times}10^{-7}$ & $2.14{\times}10^{-7}$ & 152 & 2690 \\
Fan-in Muon ($3{\times}10^{-5}$) & $7.98{\times}10^{-7}$ & $6.84{\times}10^{-7}$ & 277 & 2773 \\
RMS-matched Muon ($10^{-6}$) & $1.81{\times}10^{-7}$ & $1.77{\times}10^{-7}$ & 280 & 2709 \\
\bottomrule
\end{tabular}
\caption{Matched five-update scale and timing diagnostic. RMS columns average
the applied parameter update over five optimizer steps; optimizer latency is
the step median, and training time is cumulative through update five.}
\label{tab:empirical-update-rms}
\end{table}

Standard high-rate Muon applies $3.53\times$ AdamW's hidden-matrix RMS and
$3.20\times$ its total RMS. The matched convention brings these ratios to
$0.80$ and $0.83$, respectively. Newton--Schulz increases median optimizer-step
latency by about $1.8\times$, while cumulative five-update training time differs
by at most $3.1\%$ because rollout and policy evaluation dominate the step.
The table isolates scale matching and local optimizer overhead.

\paragraph{Full-budget magnitude control.}
We next evaluate the RMS-matched convention for the full training budget in
a matched GiGPO comparison. Table~\ref{tab:rmsmatched-outcome}
contrasts it with the fixed AdamW reference and standard fan-in Muon recipe.

\begin{table}[h]
\centering
\small
\setlength{\tabcolsep}{8pt}
\begin{tabular}{@{}lrrr@{}}
\toprule
Recipe & Final & Late & Norm. AUC \\
\midrule
AdamW ($10^{-6}$) & 0.430 & 0.384 & 0.153 \\
Fan-in Muon ($3{\times}10^{-5}$) & \textbf{0.727} & \textbf{0.656} & \textbf{0.192} \\
RMS-matched Muon ($10^{-6}$) & 0.297 & 0.255 & 0.096 \\
\bottomrule
\end{tabular}
\caption{Full-budget magnitude sensitivity at 0.5B GiGPO. RMS matching removes
the late-success advantage of the standard fan-in recipe.}
\label{tab:rmsmatched-outcome}
\end{table}

The magnitude control shows that the standard recipe changes the applied-update
scale, not only the nominal learning-rate label. Because the fan-in and
RMS-matched conventions also change layerwise scales, this control is a
magnitude and implementation-convention check rather than a pure spectral-shape
ablation. At this operating point, comparable global RMS alone does not recover
the fan-in Muon gain. In a matched full-budget measurement, mean hidden-matrix
applied-update RMS is $1.99\times10^{-7}$ for AdamW and
$8.05\times10^{-7}$ for fan-in Muon; the corresponding total RMS values are
$1.77\times10^{-7}$ and $6.90\times10^{-7}$. Median optimizer-step latency is
152 versus 276 ms, while cumulative training time is 38,300 versus 38,753 s.
Thus the local $1.82\times$ optimizer overhead changes end-to-end training
time by only $1.2\%$.

\section{C. Additional Experimental Results}
\label{supp:additional-results}

This section complements the main results with credit-control ablations,
matched comparisons, task-level breakdowns, and optimization diagnostics.
Unless noted otherwise, faint trajectories show raw checkpoints or traces, and
heavy trajectories show five-point trailing means computed within each curve.
Late success is the mean over evaluations at updates $175$--$200$; normalized
AUC is the trapezoidal area over updates $0$--$200$, divided by $200$.

Matched comparisons share the model, estimator, data, rollout budget, training
schedule, environment initialization, task split, evaluation set, and all
non-optimizer settings, with five distinct random seeds per estimator--rate
cell.

\subsection{C.1 Credit-Control Interaction}
\label{supp:credit-control}

\ifdefined\SUPPLEMENTARYDOC
Figure~\ref{fig:omega-factorial-supp} shows the matched trajectories.
\else
Figure~\ref{fig:omega-factorial} in the main paper shows the matched
trajectories.
\fi
Across the matched factorial comparisons, enabling the GiGPO step term
raises late success for both AdamW and Muon. The Muon late-success effect is
positive in all $\omega=1$ comparisons and in two of three $\omega=0$
comparisons, with more variable normalized-AUC effects. These results separate
the optimizer and credit effects, with the interaction varying across
comparisons.

\begin{table}[H]
\centering
\small
\setlength{\tabcolsep}{6pt}
\begin{tabular}{@{}lrrrr@{}}
\toprule
& \multicolumn{2}{c}{$\omega=0$} & \multicolumn{2}{c}{$\omega=1$} \\
\cmidrule(lr){2-3}\cmidrule(l){4-5}
Comparison & $\Delta$ late & $\Delta$ AUC & $\Delta$ late & $\Delta$ AUC \\
\midrule
1 & $+0.220$ & $+0.069$ & $+0.255$ & $+0.127$ \\
2 & $+0.018$ & $+0.008$ & $+0.031$ & $-0.031$ \\
3 & $-0.009$ & $-0.012$ & $+0.272$ & $+0.038$ \\
\bottomrule
\end{tabular}
\caption{Muon ($3\times10^{-5}$) minus matched AdamW ($10^{-6}$) in the
GiGPO optimizer-by-credit factorials. Columns separate the two credit settings.}
\label{tab:omega-repetitions}
\end{table}

\ifdefined\SUPPLEMENTARYDOC
\begin{figure}[H]
\centering
\includegraphics[width=0.86\textwidth]{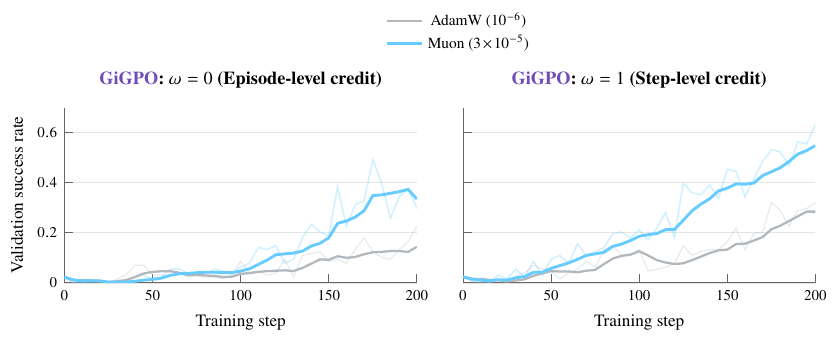}
\caption{Representative validation trajectories for the GiGPO
optimizer-by-credit factorial. Faint curves are raw checkpoints; heavy curves
are five-point trailing means. Table~\ref{tab:omega-repetitions} reports
the individual comparisons.}
\label{fig:omega-factorial-supp}
\end{figure}
\fi

\subsection{C.2 Robustness Across Matched Comparisons}
\label{supp:matched-robustness}

\ifdefined\SUPPLEMENTARYDOC
The main paper reports the GRPO and GiGPO matched effects.
\else
Figure~\ref{fig:algofamily} reports the GRPO and GiGPO matched comparisons as
optimizer effects.
\fi
Table~\ref{tab:repeated-run-effects} summarizes matched comparisons under GRPO,
GiGPO, and GraphGPO. The intervals use the $t$ distribution at the
seed level and complement the pooled direction tests.

\begin{table}[H]
\centering
\scriptsize
\setlength{\tabcolsep}{4pt}
\renewcommand{\arraystretch}{1.04}
\begin{tabular}{@{}llrrrr@{}}
\toprule
& & \multicolumn{2}{c}{Late success} & \multicolumn{2}{c}{Norm. AUC} \\
\cmidrule(lr){3-4}\cmidrule(l){5-6}
Estimator & Muon lr & $\Delta$ & 95\% CI & $\Delta$ & 95\% CI \\
\midrule
GRPO & $10^{-5}$ & $+0.095$ & $[0.009,0.182]$ & $+0.043$ & $[-0.002,0.088]$ \\
 & $3{\times}10^{-5}$ & $+0.110$ & $[-0.034,0.254]$ & $+0.045$ & $[-0.021,0.112]$ \\
\addlinespace[1pt]
GiGPO & $10^{-5}$ & $+0.027$ & $[-0.186,0.239]$ & $+0.029$ & $[-0.070,0.128]$ \\
 & $3{\times}10^{-5}$ & $+0.237$ & $[0.035,0.439]$ & $+0.073$ & $[-0.043,0.189]$ \\
\addlinespace[1pt]
GraphGPO & $10^{-5}$ & $+0.123$ & $[0.075,0.170]$ & $+0.152$ & $[0.073,0.230]$ \\
 & $3{\times}10^{-5}$ & $+0.110$ & $[0.034,0.187]$ & $+0.150$ & $[0.085,0.215]$ \\
\bottomrule
\end{tabular}
\renewcommand{\arraystretch}{1.0}
\caption{Effects across matched comparisons at 0.5B
relative to AdamW $10^{-6}$ (mean and two-sided 95\% $t$ interval).}
\label{tab:repeated-run-effects}
\end{table}

Removing the matched group with the largest high-rate late-success difference
leaves the late-success direction test significant. Its exact two-sided
sign-test value is $p=.0018$, and the result remains significant after Holm
correction ($p_{\mathrm{adj}}=.0073$).

As a sensitivity check, we discard the pairing assignments and permute optimizer
labels within each estimator stratum while preserving the reported AdamW and
Muon group sizes. Exhaustive enumeration gives 16,003,008 assignments per test.
Table~\ref{tab:stratified-permutation} shows that the high-rate result remains
significant under this alternative analysis for both metrics. The permutation
test uses effect magnitudes, whereas the sign test in the main paper uses paired
directions, so the two analyses answer different questions for AUC.

\begin{table}[H]
\centering
\small
\setlength{\tabcolsep}{7pt}
\renewcommand{\arraystretch}{1.04}
\begin{tabular}{@{}llrrr@{}}
\toprule
Muon lr & Metric & Mean diff. & Exact $p$ & Holm $p$ \\
\midrule
$10^{-5}$ & Late success & 0.082 & 0.0050 & 0.0050 \\
 & Norm. AUC & 0.074 & 0.0002 & 0.0003 \\
\addlinespace[1pt]
$3{\times}10^{-5}$ & Late success & 0.152 & $2.6{\times}10^{-5}$ & $7.9{\times}10^{-5}$ \\
 & Norm. AUC & 0.089 & $1.8{\times}10^{-5}$ & $7.3{\times}10^{-5}$ \\
\bottomrule
\end{tabular}
\renewcommand{\arraystretch}{1.0}
\caption{Exact estimator-stratified permutation sensitivity after discarding
the pairing assignments. Mean differences pool outcomes across strata;
Holm adjustment covers the same four rate--metric tests as the paired analysis.}
\label{tab:stratified-permutation}
\end{table}

\begin{figure}[H]
\centering
\includegraphics[width=0.88\textwidth]{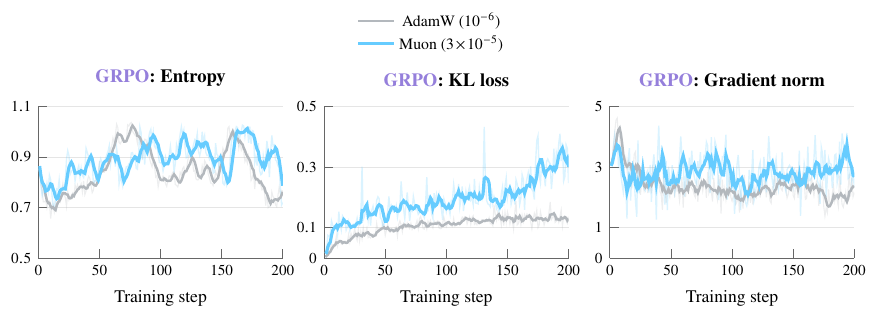}
\caption{Training dynamics for the main GRPO AdamW and Muon
($3\times10^{-5}$) comparison. Faint lines are raw per-step traces; heavy lines
are five-point trailing means. Entropy is response-token policy entropy; Muon moves
farther while both gradients remain within the plotted range.}
\label{fig:grpo-dynamics}
\end{figure}

Figure~\ref{fig:grpo-dynamics} shows stable GRPO training through step $200$.
Muon moves farther from the reference policy while remaining on the same
gradient-norm scale.

\subsection{C.3 Task-Level Results}
\label{supp:task-results}

\begin{figure}[H]
\centering
\includegraphics[width=0.92\textwidth]{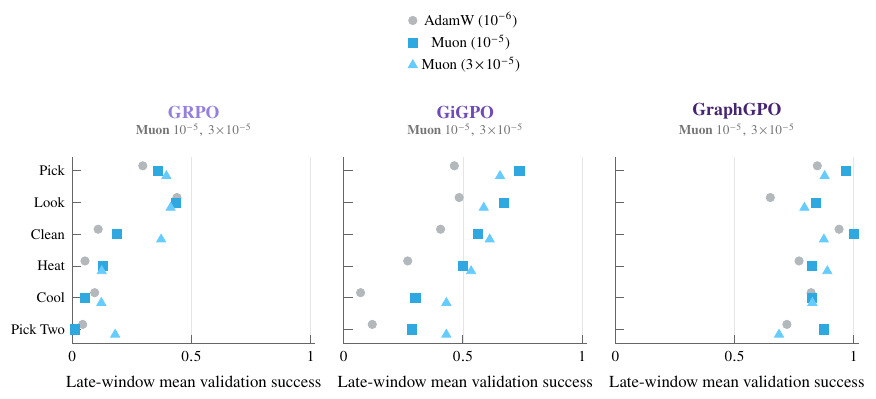}
\caption{Per-task late-window validation success in the three main
comparisons. Gray circles denote AdamW; blue squares and triangles denote the
two Muon rates. Small vertical offsets separate markers without changing the
reported success values.}
\label{fig:pertask}
\end{figure}

Figure~\ref{fig:pertask} summarizes late-window performance by task category
for all three estimators and both Muon learning rates. The differences span
multiple task categories rather than concentrating in one task family.

\subsection{C.4 Optimizer Dynamics Across Estimators}
\label{supp:optimizer-dynamics}

\begin{figure}[H]
\centering
\includegraphics[width=0.88\textwidth]{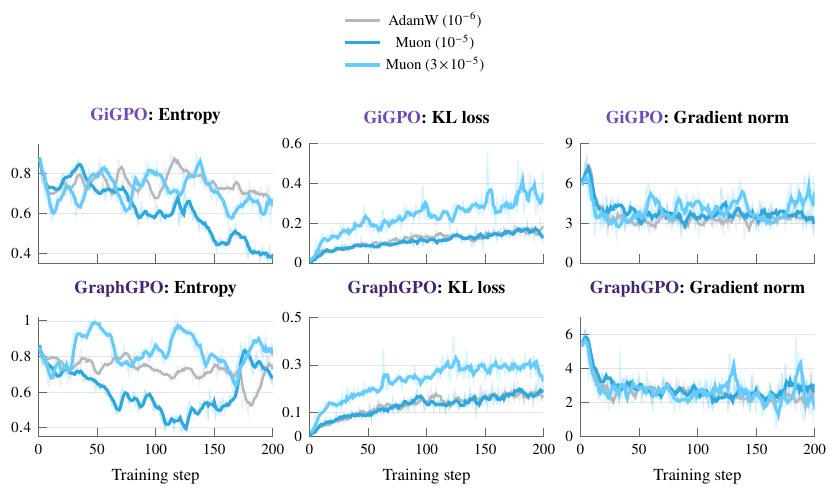}
\caption{Optimizer diagnostics under GiGPO (top) and GraphGPO (bottom).
Faint lines are raw per-step traces; heavy lines are five-point trailing means. Gradient
norms remain comparable, while KL movement depends on estimator and Muon rate.}
\label{fig:dynamics}
\end{figure}

Figure~\ref{fig:dynamics} compares optimizer dynamics across estimators. The
clearest separation lies in policy movement rather than overall gradient
magnitude: KL trajectories vary across configurations, whereas gradient norms
remain on a comparable scale.

\subsection{C.5 Scale Extensions}
\label{supp:scale-extensions}

\begin{figure}[H]
\centering
\includegraphics[width=0.72\textwidth]{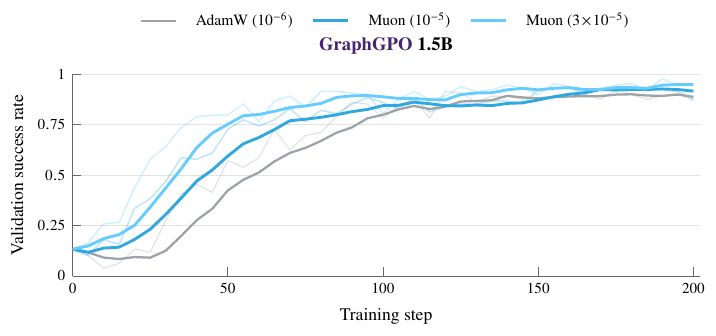}
\caption{Validation success for the matched 1.5B GraphGPO comparison. Faint
curves are raw checkpoints; heavy curves are five-point trailing means.
Both Muon rates learn faster than AdamW; the higher Muon rate also retains the
largest late-window success.}
\label{fig:graphgpo-q15}
\end{figure}

The matched 1.5B GraphGPO comparison preserves the optimizer separation in both
late-window success and normalized AUC (Figure~\ref{fig:graphgpo-q15}).
Muon at $3\times10^{-5}$ gives the strongest trajectory, raising normalized
AUC from $0.675$ to $0.805$ and late-window success from $0.893$ to $0.948$.

\begin{table}[H]
\centering
\small
\setlength{\tabcolsep}{8pt}
\begin{tabular}{@{}llcc@{}}
\toprule
Estimator & Optimizer (lr) & Late success & Norm. AUC \\
\midrule
\multirow{3}{*}{GRPO}
 & AdamW ($10^{-6}$) & 0.598 & 0.398 \\
 & Muon ($10^{-5}$) & \textbf{0.814} & 0.529 \\
 & Muon ($3{\times}10^{-5}$) & 0.813 & \textbf{0.574} \\
\midrule
\multirow{3}{*}{GiGPO}
 & AdamW ($10^{-6}$) & \textbf{0.926} & 0.635 \\
 & Muon ($10^{-5}$) & 0.915 & 0.683 \\
 & Muon ($3{\times}10^{-5}$) & 0.910 & \textbf{0.708} \\
\midrule
\multirow{3}{*}{GraphGPO}
 & AdamW ($10^{-6}$) & 0.893 & 0.675 \\
 & Muon ($10^{-5}$) & 0.915 & 0.740 \\
 & Muon ($3{\times}10^{-5}$) & \textbf{0.948} & \textbf{0.805} \\
\bottomrule
\end{tabular}
\caption{Matched 1.5B scale extensions. Bold marks the largest value within an
estimator and metric. Muon has higher normalized AUC across the evaluated
estimators, while late-window rankings reflect different degrees of saturation.}
\label{tab:q15-summary}
\end{table}

For the same 1.5B configurations, we record lightweight online probes over six
uniformly selected hidden matrices (Table~\ref{tab:graphgpo-q15-diagnostics}).
The applied-update row participation ratio is similar across optimizers, and
the temporal cosine between one update direction and the next training batch's
gradient is small across configurations. These columns summarize local update
geometry; row PR is a sketch-based row statistic and differs from the
full-matrix normalized stable rank reported in the main paper.

\begin{table}[H]
\centering
\small
\setlength{\tabcolsep}{6pt}
\begin{tabular}{@{}lrrrr@{}}
\toprule
Configuration & Late & Norm. AUC & Row PR & Next-grad cosine \\
\midrule
AdamW ($10^{-6}$) & 0.893 & 0.675 & 0.821 & 0.010 \\
Muon ($10^{-5}$) & 0.915 & 0.740 & 0.793 & 0.012 \\
Muon ($3{\times}10^{-5}$) & \textbf{0.948} & \textbf{0.805} & 0.799 & 0.006 \\
\bottomrule
\end{tabular}
\caption{1.5B GraphGPO outcome and online diagnostics. Row PR is the
equal-layer mean participation ratio of the applied optimizer update;
next-grad cosine compares that direction with the subsequent training-batch
gradient. Diagnostic columns average the training trajectory.}
\label{tab:graphgpo-q15-diagnostics}
\end{table}

At 3B, the GraphGPO comparison includes both the default AdamW
$10^{-6}$ rate and a tuned $3\times10^{-6}$ control. High-rate Muon reaches
$0.75$ success in $30$ updates, compared with $65$ for default AdamW and $40$
for tuned AdamW; its normalized AUC is $0.856$, versus $0.853$ for tuned AdamW.
The full trajectories illustrate this early-progress boundary case.

\paragraph{GiGPO scale trajectories.}

\begin{figure}[H]
\centering
\includegraphics[width=0.86\textwidth]{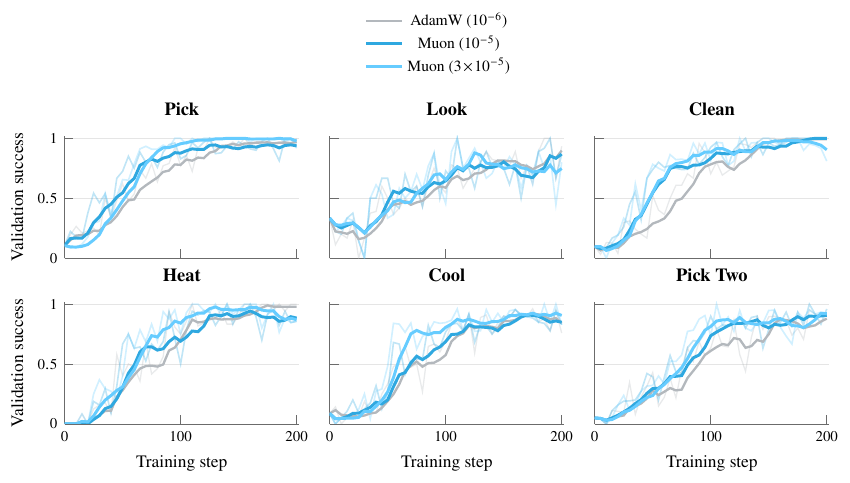}
\caption{Per-task validation trajectories for the 1.5B GiGPO comparison.
Faint curves show raw task checkpoints; heavier curves show five-point
trailing means for the same task and optimizer. All configurations approach
high success, while Muon reaches that regime earlier across most task
categories.}
\label{fig:q15-pertask}
\end{figure}

\begin{figure}[H]
\centering
\includegraphics[width=0.86\textwidth]{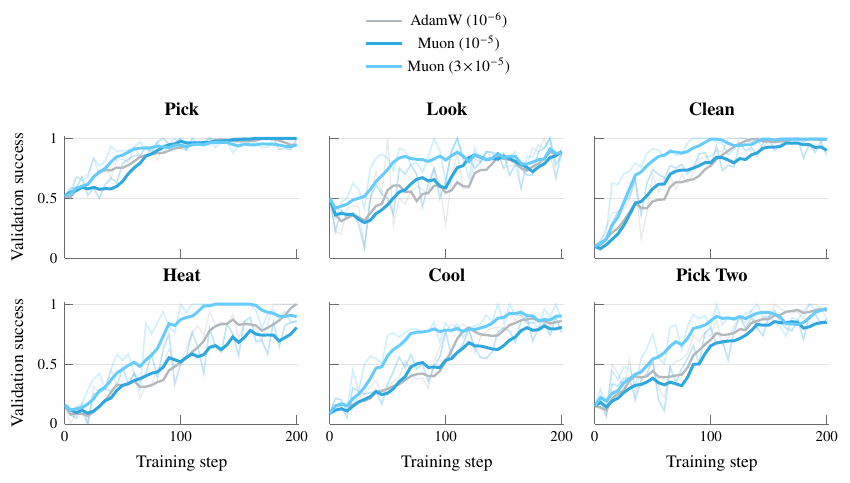}
\caption{Per-task validation trajectories for the 3B GiGPO comparison. Faint
curves show raw task checkpoints; heavier curves show five-point trailing
means for the same task and optimizer. Muon at $3\times10^{-5}$ learns fastest
across most categories; the lower Muon rate finishes below the AdamW and
higher-rate Muon configurations.}
\label{fig:q3-pertask}
\end{figure}

\begin{figure}[H]
\centering
\includegraphics[width=0.78\textwidth]{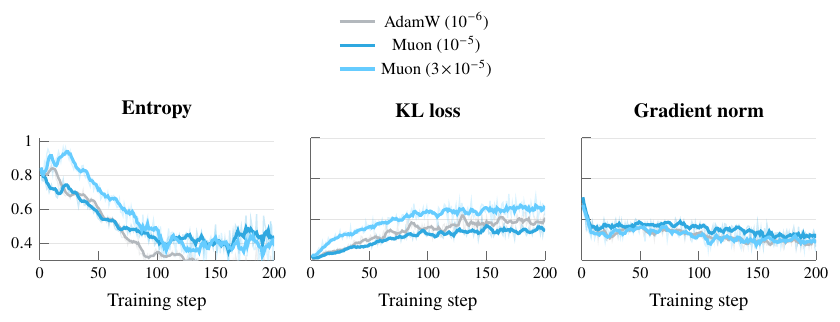}
\caption{Training diagnostics for the 1.5B GiGPO comparison. Faint lines are
raw per-step traces; heavy lines are five-point trailing means. Muon improves
faster without a distinct gradient-norm shift. Entropy is
response-token policy entropy.}
\label{fig:q15-dynamics}
\end{figure}

Each GiGPO scale comparison uses a matched AdamW--Muon configuration set over
the same fixed training horizon. Figures~\ref{fig:q15-pertask}
and~\ref{fig:q3-pertask} show all six task categories at 1.5B and 3B,
respectively. The aggregate differences appear mainly in the speed of learning
rather than in the saturated late window. Figure~\ref{fig:q15-dynamics}
provides the corresponding 1.5B step-level diagnostics.

\FloatBarrier
\subsection{C.6 WebShop Transfer Case}
\label{supp:webshop}

\begin{figure}[H]
\centering
\includegraphics[width=0.78\textwidth]{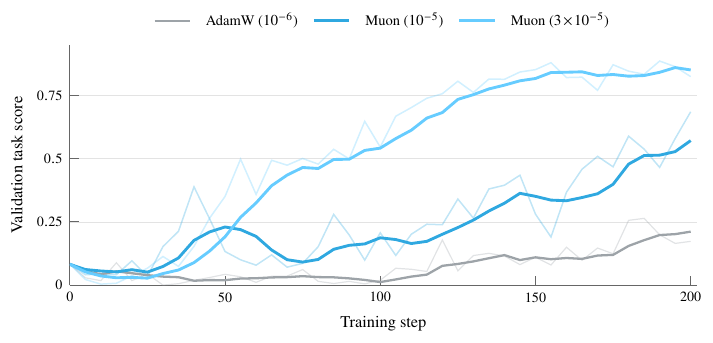}
\caption{WebShop partial-task-score trajectories for the matched 0.5B GiGPO
comparison. Faint curves are raw five-update evaluations; heavy curves are
five-point trailing means. High-rate Muon has the strongest sustained
trajectory.}
\label{fig:webshop-stress}
\end{figure}

Figure~\ref{fig:webshop-stress} shows the continuous partial task score for a
matched 0.5B GiGPO optimizer comparison on WebShop. High-rate Muon crosses task
score $0.5$ by update 75 and stays above this threshold through the end of
training; lower-rate Muon improves more gradually, while AdamW remains near its
initial score range.

\begin{table}[H]
\centering
\small
\setlength{\tabcolsep}{7pt}
\begin{tabular}{@{}lrrrr@{}}
\toprule
Configuration & Task late & Task AUC & Success late & Success AUC \\
\midrule
AdamW ($10^{-6}$) & 0.197 & 0.078 & 0.008 & 0.005 \\
Muon ($10^{-5}$) & 0.554 & 0.256 & 0.168 & 0.047 \\
Muon ($3{\times}10^{-5}$) & \textbf{0.854} & \textbf{0.548} & \textbf{0.689} & \textbf{0.305} \\
\bottomrule
\end{tabular}
\caption{WebShop outcomes for the matched comparison. Task score provides partial
credit; success requires exact task completion.}
\label{tab:webshop-summary}
\end{table}

High-rate Muon reaches peak exact success $0.766$ and finishes at
$0.688$, compared with AdamW's final $0.023$. The continuous task score
reports partial progress alongside exact completion.

\FloatBarrier
\subsection{C.7 Applied-Update Spectrum at 1.5B}
\label{supp:applied-spectrum}

The main paper reports a matched 1.5B GRPO comparison. Muon
at $10^{-5}$ and $3\times10^{-5}$ raises late-window success from $0.598$ to
$0.814$ and $0.813$, respectively, and normalized AUC from $0.398$ to $0.529$
and $0.574$. For a probed matrix $X\in\mathbb R^{m\times n}$, we use normalized
stable rank
\(
r_{\mathrm{st}}(X)=\|X\|_F^2/
(\|X\|_2^2\min\{m,n\})
\), which lies in $(0,1]$ and increases as the singular spectrum flattens.
At each update, we aggregate this statistic over the probed hidden-weight
matrices before and after the optimizer transformation, using the same fixed
matrix set across configurations and an equal-weight mean across matrices.
Averaged over training, the pre-transform first-moment statistic is
$0.0051$ for AdamW and approximately
$0.0014$ for the Muon configurations. The applied-update statistic is $0.015$ for
AdamW, versus $0.585$ and $0.635$ for Muon. Thus the diagnostic confirms
spectral flattening for the probed applied updates.

\FloatBarrier
\section{D. Spectral--Credit Compatibility Hypothesis}
\label{supp:credit-hypothesis}

This section develops an explanatory compatibility view suggested by the
spectral diagnostics: Muon's flattened update is most useful when the promoted
directions carry reliable policy-improvement signal.

For estimator $a$, write its stochastic matrix gradient as
\begin{equation}
\widehat G_a=G_a^\star+E_a,
\end{equation}
where $G_a^\star$ is the population direction and $E_a$ is credit and sampling
noise. Let $\Pi_{\mathrm{tail}}$ project onto weak directions in a common
raw-gradient basis and define
\begin{equation}
\mathcal R_a^{\mathrm{tail}}
=\frac{\|\Pi_{\mathrm{tail}}G_a^\star\|_F^2}
{\mathbb E\|\Pi_{\mathrm{tail}}E_a\|_F^2+\varepsilon}.
\label{eq:tail-reliability}
\end{equation}
In a fixed-basis model with singular-direction signal $s_i>0$ and Gaussian
noise $\epsilon_i\sim\mathcal N(0,\sigma_i^2)$, an ideal polar update has
expected alignment
\begin{equation}
\mathbb E\langle\mathcal P(\widehat G),G^\star\rangle_F
=\sum_i s_i\!\left[2\Phi(s_i/\sigma_i)-1\right].
\label{eq:appendix-muon-alignment}
\end{equation}
Muon gives weak singular directions comparable magnitude, so the transform is
most useful when those directions are sufficiently reliable. The quantity
$\mathcal R_a^{\mathrm{tail}}$ summarizes this reliability condition for
spectral flattening.

\FloatBarrier
\section{E. Learning-Rate Controls and Stress Diagnostics}
\label{supp:rate-and-failures}

\subsection{E.1 AdamW Learning-Rate Controls}
\label{supp:rate-controls}

Table~\ref{tab:adamw-rate-controls} summarizes the 0.5B AdamW
rate controls. GiGPO includes a competitive $3\times10^{-6}$ outcome and a
$3\times10^{-5}$ stress setting. Under GRPO, $10^{-6}$ remains strongest and
$10^{-5}$ has zero late success.

\begin{table}[H]
\centering
\small
\setlength{\tabcolsep}{5.5pt}
\begin{tabular}{@{}llcccc@{}}
\toprule
Estimator & AdamW lr & \multicolumn{2}{c}{Late success} &
\multicolumn{2}{c}{Norm. AUC} \\
\cmidrule(lr){3-4}\cmidrule(l){5-6}
 & & Mean & Max & Mean & Max \\
\midrule
\multirow{5}{*}{GiGPO}
 & $10^{-6}$           & \textbf{0.345} & 0.448 & \textbf{0.140} & 0.180 \\
 & $3{\times}10^{-6}$ & 0.282 & \textbf{0.628} & 0.124 & \textbf{0.290} \\
 & $5{\times}10^{-6}$ & 0.073 & 0.099 & 0.025 & 0.035 \\
 & $10^{-5}$           & 0.000 & 0.000 & 0.000 & 0.000 \\
 & $3{\times}10^{-5}$ & 0.000 & 0.000 & 0.000 & 0.000 \\
\midrule
\multirow{4}{*}{GRPO}
 & $10^{-6}$           & \textbf{0.105} & \textbf{0.161} & \textbf{0.051} & \textbf{0.078} \\
 & $3{\times}10^{-6}$ & 0.087 & 0.124 & 0.033 & 0.049 \\
 & $5{\times}10^{-6}$ & 0.051 & 0.062 & 0.020 & 0.028 \\
 & $10^{-5}$           & 0.000 & 0.000 & 0.001 & 0.001 \\
\bottomrule
\end{tabular}
\caption{AdamW learning-rate controls at 0.5B. Means and maxima summarize the
reported outcomes; $3\times10^{-5}$ is GiGPO-only.}
\label{tab:adamw-rate-controls}
\end{table}

\paragraph{Intermediate-rate dispersion.}
GiGPO AdamW at $3\times10^{-6}$ spans late success from $0.026$ to $0.628$,
with a mean below the $10^{-6}$ reference even though the best individual
control is stronger. Higher AdamW rates fall outside the stable range under the
shared regularization; the intermediate rate mainly increases dispersion rather
than average performance.

\subsection{E.2 High-Rate Stress Diagnostics}
\label{supp:failure-diagnostics}

\begin{figure}[H]
\centering
\includegraphics[width=0.72\textwidth]{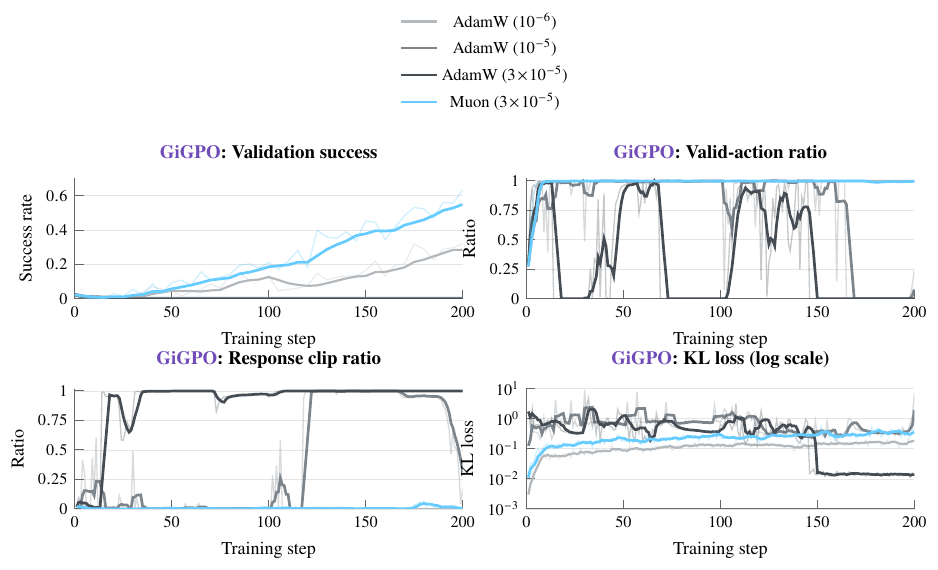}
\caption{GiGPO high-rate diagnostics. Faint curves show raw traces; heavy
curves show five-point trailing means, with KL on a logarithmic axis.}
\label{fig:lr-stress}
\end{figure}

Figure~\ref{fig:lr-stress} summarizes the high-rate setting under the shared KL
and clipping recipe. AdamW loses valid behavior at $3\times10^{-5}$ and shows
late KL excursions at $10^{-5}$; Muon at
$3\times10^{-5}$ retains valid behavior and improves success. With the KL
coefficient fixed, these trajectories show high-rate behavior under the same
regularization setting.

\ifdefined\SUPPLEMENTARYDOC
\enlargethispage{4\baselineskip}
\fi

\end{document}